\documentclass[]{youtu}

\usepackage{mathpazo}

\usepackage{graphicx}
\usepackage[numbers]{natbib}
\usepackage{times}
\usepackage[utf8]{inputenc} 
\usepackage[T1]{fontenc}    

\usepackage{url}          
\usepackage{booktabs}      
\usepackage{amsfonts}      
\usepackage{nicefrac}      
\usepackage{microtype}     
\usepackage{epsfig}
\usepackage{float}
\usepackage{multicol}
\usepackage{multirow}
\usepackage{amssymb}
\usepackage{amsmath}
\usepackage{wrapfig}

\usepackage{xspace}
\usepackage{color}
\usepackage{colortbl}
\usepackage[dvipsnames, svgnames, x11names, table]{xcolor}
\usepackage[misc]{ifsym}
\usepackage{makecell}
\usepackage{soul}
\usepackage{bm}

\usepackage{subcaption, overpic, textpos}

\usepackage{paralist}
\usepackage{tabularx}

\usepackage{algorithm}
\usepackage{algorithmic}
\usepackage{adjustbox}
\usepackage{pifont}
\usepackage{longtable} 
\usepackage{titletoc}

\usepackage[capitalize]{cleveref}
\crefname{section}{Sec.}{Secs.}
\Crefname{section}{Section}{Sections}
\Crefname{table}{Table}{Tables}
\crefname{table}{Tab.}{Tabs.}

\newlength\savewidth

\renewcommand{\paragraph}[1]{\vspace{1.25mm}\noindent\textbf{#1}}

\newcommand{\cmark}{\ding{52}\xspace}%
\newcommand{\xmark}{\ding{56}\xspace}%

\definecolor{lightgray}{rgb}{0.8, 0.8, 0.8}
\definecolor{lgray}{rgb}{0.66, 0.66, 0.66}
\definecolor{whit_tab}{RGB}{255, 255, 255}
\definecolor{gray_tab}{RGB}{246, 246, 246}
\definecolor{oran_tab}{RGB}{252, 242, 237}
\definecolor{blue_tab}{RGB}{227, 240, 251}
\definecolor{lblu_tab}{RGB}{225, 235, 246}
\definecolor{orange_vitad}{RGB}{222, 131, 68}
\definecolor{blue_vitad}{RGB}{106, 153, 208}

\definecolor{trajectory_green}{RGB}{126, 171, 85}
\definecolor{trajectory_yellow}{RGB}{245, 194, 66}

\def\method{JAVEdit}
\def\dataset{JAVEdit-100k}
\def\model{JAVEdit}
\def\benchmark{JAVEditBench}

\title{\textsc{\method}: Joint Audio-Visual Instruction-Guided Video Editing with Agentic Data Curation}

\author{
Yinan Chen$^{1*}$\quad
Chuming Lin$^{2*}$\quad
Zhennan Chen$^3$\quad
Yuxiang Zeng$^4$\quad
Junwei Zhu$^2$ \\
Yali Bi$^1$\quad
Xijie Huang$^5$\quad
Chengming Xu$^2$\quad
Donghao Luo$^2$\quad
Zhucun Xue$^1$ \\
Xiaobin Hu$^6$\quad
Chengjie Wang$^2$\quad
Yong Liu$^1$\quad
Jiangning Zhang$^{1,2\dagger}$\quad
Shuicheng Yan$^6$
}
\affiliation{
$^1$Zhejiang University \quad
$^2$Tencent Youtu Lab \quad
$^3$Nanjing University \\
$^4$University of Auckland \quad
$^5$Fudan University \quad
$^6$National University of Singapore \\
}

\date{May 30, 2026}
\correspondence{yinan.chen@zju.edu.cn} 
\sourcecode{https://github.com/RyanChenYN/JAVEdit}
\data{https://huggingface.co/datasets/Coraxor/JAVEdit-100k}
\project{https://ryanchenyn.github.io/projects/JAVEdit}

\begin{document}

\abstract{While instruction-based video editing has seen significant progress, joint audio-visual editing remains constrained by the absence of dedicated datasets and benchmarks. To bridge this gap, we present \textbf{\dataset}, the first large-scale, high-quality dataset tailored for instruction-guided joint audio-visual editing. Focusing on human-centric videos, \dataset\ comprises approximately 100K editing triplets spanning five distinct categories, including subject editing and speech editing. This dataset is rigorously constructed via four meticulously designed generation pipelines, seamlessly paired with an agent-in-the-loop quality control mechanism. Furthermore, to address the lack of standardized evaluation within the field, we introduce \textbf{\benchmark}, a comprehensive benchmark featuring curated source videos and human-aligned instructions across all editing categories. Finally, we propose \textbf{\model}, a pioneering baseline model for instruction-guided joint audio-visual editing. Experiments show that \model\ outperforms all baselines on five of six evaluation metrics. All data, code, and model weights will be publicly released.}

\maketitle

{
\let\thefootnote\relax
\footnotetext{$^*$ indicates equal contributions. $^\dagger$ indicates corresponding author. This work was done when Yinan Chen was an intern at Tencent Youtu Lab.}\quad
}

\section{Introduction}
\label{sec:intro}

Recent advancements in instruction-guided video editing~\cite{InsViE-1M} have demonstrated remarkable capabilities, largely propelled by the emergence of high-quality training datasets~\cite{OpenVE-3M}. However, research on models and datasets for joint audio-visual editing remains conspicuously insufficient. This scarcity primarily stems from the challenge of maintaining strict spatiotemporal and semantic alignment between visual and audio modalities during the generation process. Furthermore, synthesizing such complex data typically involves cascading multiple specialized generative models, inevitably leading to cross-stage error accumulation. While recent pipelines~\cite{Ditto} attempt to mitigate this by relying on a "human-in-the-loop" paradigm, this manual inspection and refinement process creates a bottleneck, fundamentally prohibiting the scalable construction of large, diverse datasets.

\begin{figure*}[t]
    \centering
    \includegraphics[width=1.0\linewidth]{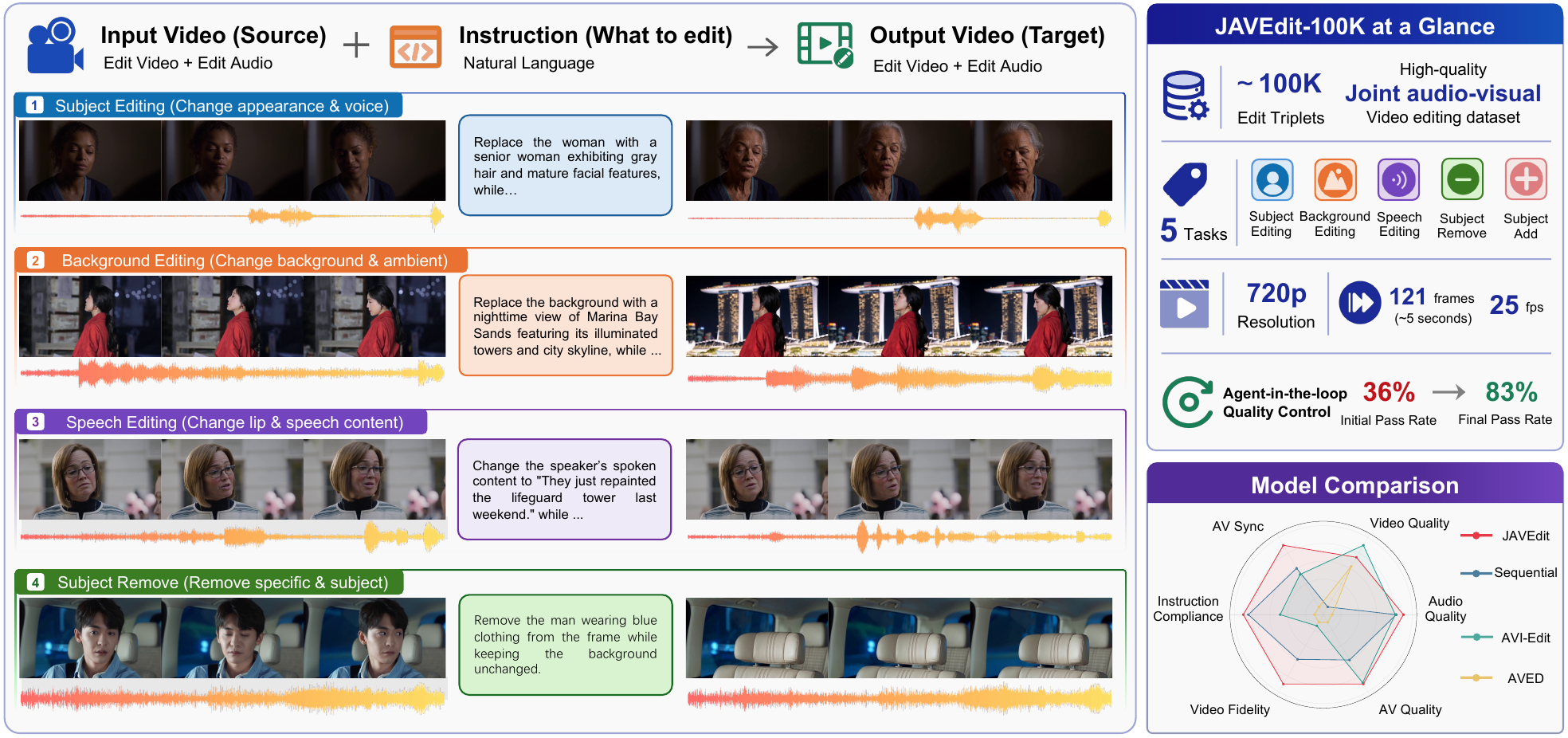}
    \vspace{-1em}
    \caption{\textbf{Overview of \method.}
    We present three components: \textbf{\dataset}, a 100K-scale dataset with Agent-in-the-loop curation;
    \textbf{\model}, a joint audio-visual editing model;
    and \textbf{\benchmark}, a benchmark with fine-grained cross-modal metrics.}
    \label{fig:teaser}
\end{figure*}

Consequently, existing video editing datasets like InsViE-1M~\cite{InsViE-1M}, Ditto~\cite{Ditto}, and OpenVE-3M~\cite{OpenVE-3M} focus exclusively on visual transformations. While a few pioneering works in joint audio-visual editing (AVED~\cite{AVED}, AVEdit~\cite{AVEdit}, AVIEdit~\cite{AVIEdit}) have emerged, they predominantly rely on a cumbersome "source-target prompt" paradigm rather than natural, user-friendly language instructions. Moreover, these datasets are largely confined to superficial attribute modifications (e.g., global style transfer) and fail to encompass structural transformations, such as subject addition/removal or fine-grained speech editing, which are essential for human-centric video editing.

To this end, we introduce \dataset, the first large-scale, high-quality dataset tailored for instruction-guided joint audio-visual editing, as illustrated in Figure~\ref{fig:teaser}. Focusing on human-centric scenarios, \dataset\ comprises approximately 100K meticulously curated editing pairs across five distinct categories. To overcome the aforementioned alignment challenges at scale, this dataset is synthesized via an automated generation pipeline empowered by a novel Agent-in-the-loop quality control mechanism, which rigorously filters and refines data to ensure strict cross-modal synchronization and instruction compliance. Furthermore, we propose \benchmark, a comprehensive benchmark featuring curated source videos and human-aligned instructions. Beyond conventional visual metrics, \benchmark\ introduces fine-grained criteria specifically designed to jointly evaluate visual–audio quality, instruction compliance, and video fidelity. Finally, we present \model, a strong baseline for the joint audio-visual editing task, which achieves excellent performance across various quantitative and qualitative metrics on \benchmark.

Our main contributions are as follows:
\begin{itemize}
    \item We introduce \textbf{\dataset}, the first large-scale dataset for instruction-guided joint audio-visual editing. It contains 100K high-quality, human-centric pairs across five categories, supporting complex structural and speech modifications.
    
    \item We propose a scalable automated pipeline driven by an \textbf{Agent-in-the-loop} mechanism, ensuring strict cross-modal alignment and high-quality data generation without manual bottlenecks.

    \item We establish \textbf{\benchmark}, a comprehensive evaluation benchmark that introduces fine-grained criteria specifically designed to jointly evaluate visual–audio quality, instruction compliance, and video fidelity.

    \item We provide \textbf{\model}, a strong baseline obtained by fine-tuning LTX-2.3 with LoRA on \dataset, demonstrating that our dataset directly enables effective joint audio-visual editing.

\end{itemize}
Experiments on \benchmark\ show that \model\ outperforms all baselines on five of six metrics, with a 26\% relative gain in audio-visual synchrony over the strongest sequential alternative, validating the necessity of joint modeling and agent-curated data.

\section{Curating {\dataset} Dataset}
\label{sec:dataset}

\begin{figure}[t]
    \centering
    \includegraphics[width=1.0\linewidth]{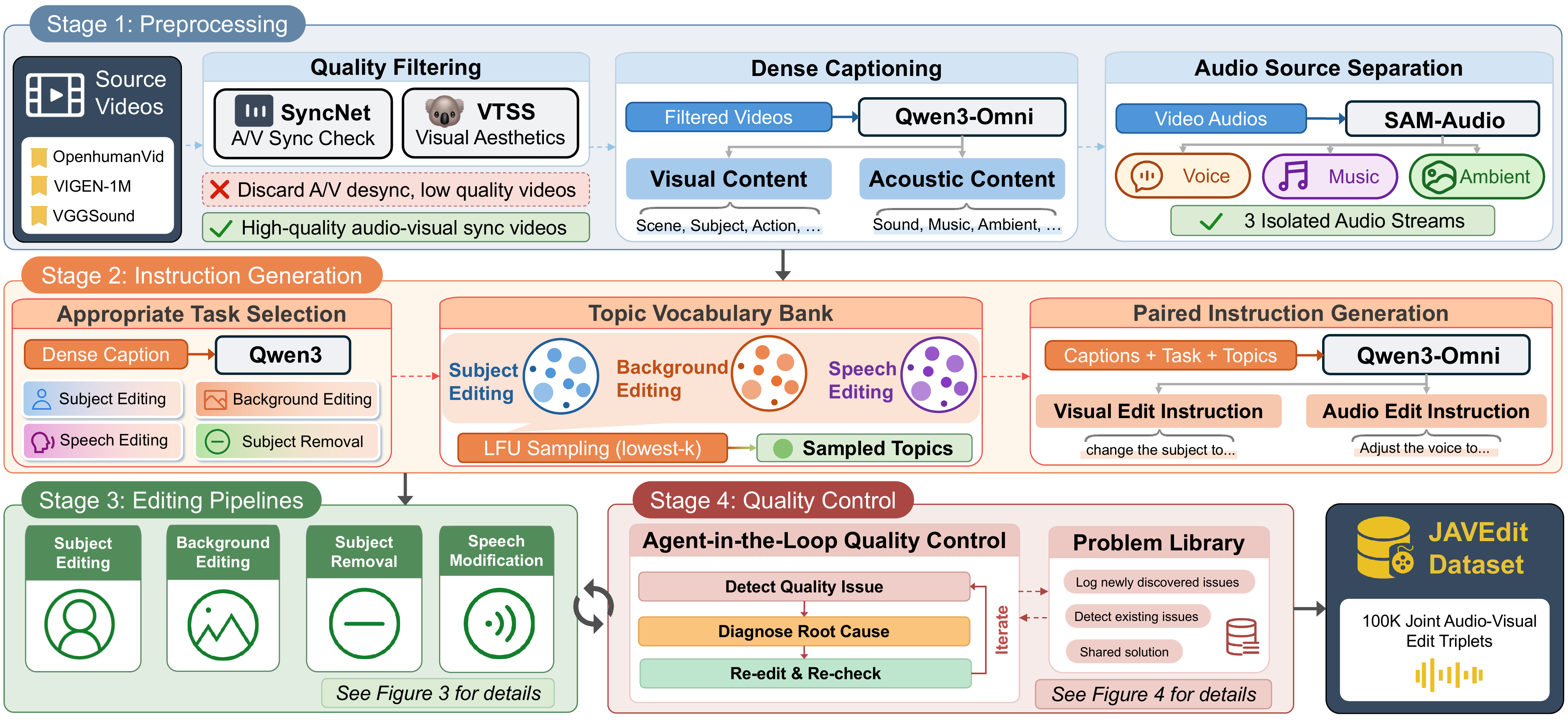}
    \vspace{-1em}
    \caption{Overview of the \dataset\ dataset construction pipeline.
    Source videos undergo preprocessing, instruction generation, category-specific editing,
    and agent-in-the-loop quality control to yield approximately 100K high-quality
    joint audio-visual editing triplets.}
    \label{fig:overview}
\end{figure}

\subsection{Task Definition and Dataset Overview}
\label{subsec:task_def}

We define \emph{instruction-guided joint audio-visual editing} as follows.
Given a source video $V$ with an accompanying audio track $A$ and a natural language instruction $I$,
the objective is to produce an edited video $V'$ with audio $A'$ that faithfully executes the specified modifications
while preserving all content unrelated to the instruction. Formally,

\begin{equation}
    (V', A') \;=\; \mathcal{T}(V, A, I),
\quad\text{s.t.}\quad
    (V', A') = (V, A) \text{ on all dimensions unspecified by } I,
    \label{eq:task_def}
\end{equation}
where $\mathcal{T}$ denotes an abstract editing operator whose concrete
instantiation as our editing model is presented in later sections.
In this work we focus on \emph{human-centric} videos, where the coupling between the
visual and audio streams is particularly tight.

The \dataset\ dataset comprises 100K editing triplets $(V, I, V')$ that span five editing categories
(Figure~\ref{fig:overview}):
\textbf{(1)~Subject Editing}, altering the appearance of the human subject
while synchronously updating the subject's voice;
\textbf{(2)~Background Editing}, altering the environment or scene,
with the ambient sound updated to match the new background;
\textbf{(3)~Subject Removal}, removing a human subject together with the associated voice;
\textbf{(4)~Subject Addition}, inserting a human subject into a scene
along with the corresponding voice;
\textbf{(5)~Speech Editing}, altering the spoken content,
with lip motion synchronized to the newly generated speech.
It is worth noting that Subject Removal and Subject Addition share a single removal pipeline
at the data level: data pairs for Subject Addition are obtained by simply reversing the input
and output of the Subject Removal pipeline, so no separate addition pipeline is required.

\begin{table}[t]
  \centering
  \scriptsize
  \setlength{\tabcolsep}{4pt}
  \caption{Comparison of \dataset\ with existing video editing datasets.
    Rows shaded in gray denote visual-only datasets that do not support joint audio-visual editing.}
  \label{tab:dataset_comparison}
  \resizebox{0.85\textwidth}{!}{%
  \begin{tabular}{llcccll}
    \toprule
    Dataset & Scale & Audio & Instruction & Agent Control & Resolution & Frame Count \\
    \midrule
    \rowcolor{gray!15}
    InsViE-1M      & $\sim$1M & \xmark & \cmark & \xmark        & 1024$\times$576   & 25 \\
    \rowcolor{gray!15}
    Se\~{n}orita-2M & $\sim$2M & \xmark & \cmark & \xmark        & 1984$\times$1280  & 100 \\
    \rowcolor{gray!15}
    Ditto-1M       & $\sim$1M    & \xmark & \cmark & \xmark        & 1280$\times$720   & 101 \\
    \rowcolor{gray!15}
    OpenVE-3M      & $\sim$3M          & \xmark & \cmark & \xmark        & 1280$\times$720   & 65--129 \\
    AVI-Edit       & $\sim$73K         & \cmark & \xmark & \xmark        & 1280$\times$720   & $\sim$240 \\
    \rowcolor{blue!8}
    \textbf{\dataset\ (Ours)} & $\sim$103K & \cmark & \cmark & \cmark & 1280$\times$720 & 121 \\
    \bottomrule
  \end{tabular}%
  }
\end{table}

\subsection{Source Video Collection and Preprocessing}
\label{subsec:preprocessing}

We collect source videos from OpenHumanVid~\cite{OpenHumanVid}, VIDGEN-1M~\cite{VIDGEN1M}, and VGGSound~\cite{VGGSound},
which together provide a large pool of raw clips on the order of millions,
and apply a three-stage preprocessing pipeline.
To ensure compatibility with downstream video generation models,
all videos are uniformly preprocessed to a resolution of $1280 \times 720$,
a total of 121 frames, and a frame rate of 25~FPS.

\noindent\textbf{Basic Quality Filtering.}

We first discard videos that lack an audio track,
and then apply audio-visual synchrony scoring using the SyncNet model from LatentSync~\cite{LatentSync}
to remove misaligned clips.
Visual aesthetic quality is subsequently assessed by the VTSS model adopted in Koala-36M~\cite{Koala36M},
and clips that fall below a predefined threshold are discarded.
After the three filtering stages, approximately 30\% of the collected videos are retained,
yielding a pool that is used to construct the final set of roughly 100K editing pairs.

\noindent\textbf{Dense Video Captioning.}
We employ Qwen3-Omni~\cite{Qwen3-Omni} to generate dense captions for each video,
covering visual content (scene, subjects, actions, and camera shots),
        acoustic content (voice characteristics, music genre, ambient sound, and atmosphere),
and temporal dynamics.
These captions serve as the semantic grounding for downstream instruction generation.

\noindent\textbf{Audio Source Separation.}
Audio source separation is a foundational component of our pipeline.
We compared several representative audio separation approaches~\cite{Mel-RoFormer,ZeroSep,SAM-Audio}
and adopted SAM-Audio for two reasons:
it delivers high-quality extraction of arbitrary semantic categories,
and it simultaneously returns the residual audio, allowing us to iteratively apply it
on the residual stream to obtain disentangled human voice, music, and ambient sound.
Using SAM-Audio, we decompose the audio of each video into up to three
disentangled streams: \emph{human voice}, \emph{music}, and \emph{ambient sound}.
These streams are stored separately, since each editing category recombines them in a different manner
(detailed in Section~\ref{subsec:pipelines}).

\begin{figure*}[t]
    \centering
    \includegraphics[width=1\linewidth]{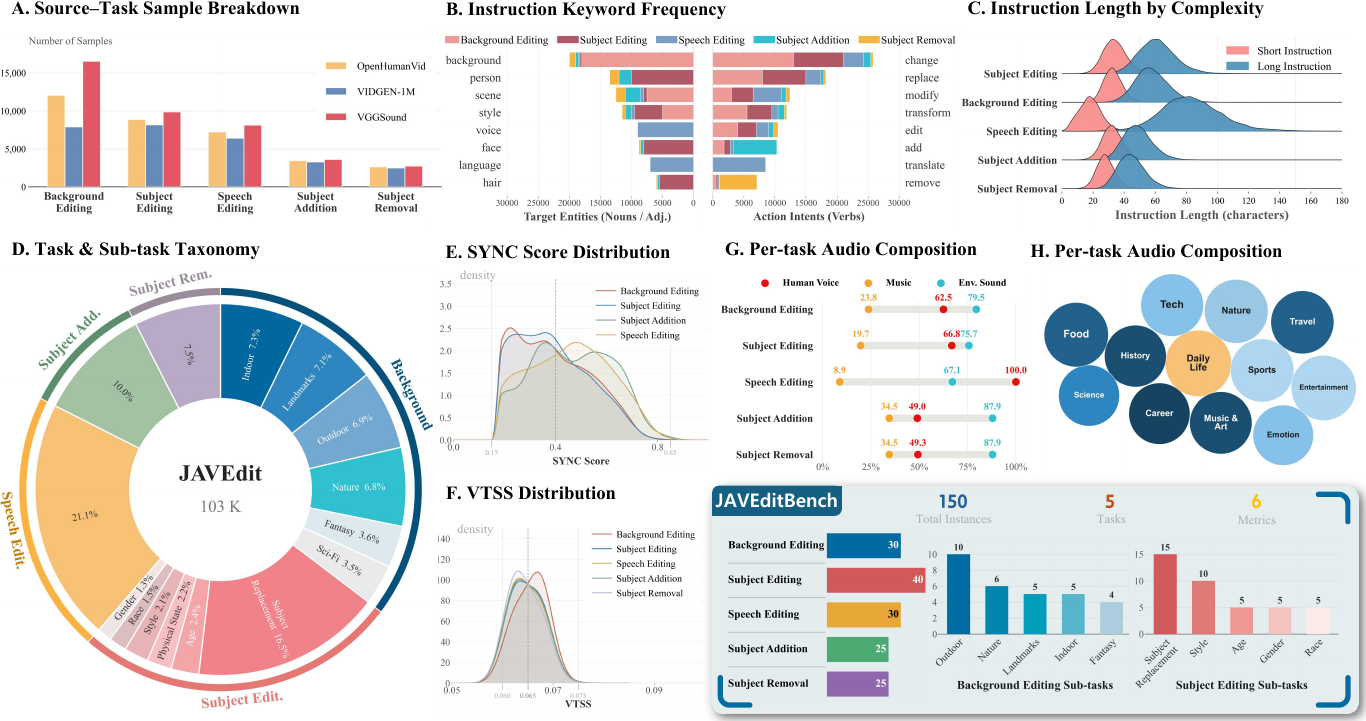}
    \vspace{-1em}
    \caption{%
      Statistics of the \dataset\ dataset.
      \textbf{(A)}~Sample counts per task broken down by source corpus.
      \textbf{(B)}~Top-8 entity (left) and action (right) keywords aggregated across all instructions.
      \textbf{(C)}~Instruction-length distributions.
      \textbf{(D)}~Task and sub-task composition.
      \textbf{(E)}~Audio-visual synchronization score distributions across four tasks (Subject Removal excluded: the edited output contains no visible face).
      \textbf{(F)}~Video quality score distributions across all five tasks.
      \textbf{(G)}~Audio component proportions per task.
      \textbf{(H)}~Topic category distribution of Speech Editing content. Please zoom in for more detail.
    }
    \label{fig:data_overview}
\end{figure*}

\subsection{Instruction Generation}
\label{subsec:instruction_gen}

\noindent\textbf{Editing Type Selection.}
For each source video, we prompt Qwen3-235B~\cite{Qwen3} with the dense caption of the video
to determine which of the four editing pipelines are applicable.
This step ensures that all generated instructions are semantically grounded;
for example, speech editing is only proposed for videos that contain a speaking subject,
and subject addition is only proposed for scenes in which an existing human subject can plausibly be removed
to construct a reversed (addition) pair.

\noindent\textbf{Topic Vocabulary Bank and Balanced Sampling.}
To promote lexical and semantic diversity, we maintain a \emph{topic vocabulary bank}
that is partitioned by editing category (e.g., appearance descriptors for subject editing,
and scene descriptors for background editing).
The vocabulary bank is initially generated by an LLM and subsequently refined through manual inspection
to remove unsuitable or ambiguous entries.
The final bank contains 6 sub-categories with 275 topic terms for subject editing,
6 sub-categories with 490 terms for background editing,
and 32 sub-categories with 1{,}230 terms for speech editing.
Subject removal does not rely on any topic description,
as the task itself is defined as removing an existing human subject from the scene,
and therefore is excluded from the vocabulary bank.
Subject addition shares the same pipeline and data pool as subject removal and likewise does not
require a dedicated vocabulary.
We further adopt a least-frequently-used sampling strategy:
at each step, Qwen3-235B is prompted to select a suitable topic from the $k$ least-sampled
candidates in the vocabulary bank, with $k$ set to 20 in practice.
This strategy prevents topic imbalance and ensures that the resulting dataset covers a wide range
of editing scenarios.

\noindent\textbf{Paired Instruction Generation.}
Given the video caption and the sampled topic, we prompt Qwen3-235B to generate
a \emph{visual editing instruction} together with a semantically consistent
\emph{audio editing instruction}, forming a paired set.
The generated instructions are required to be mutually consistent; for instance, changing the background to a rainy forest should be
paired with replacing the ambient sound with rain.

\subsection{Reliable Editing Pipelines}
\label{subsec:pipelines}

\begin{figure*}[t]
    \centering
    \includegraphics[width=1.0\linewidth]{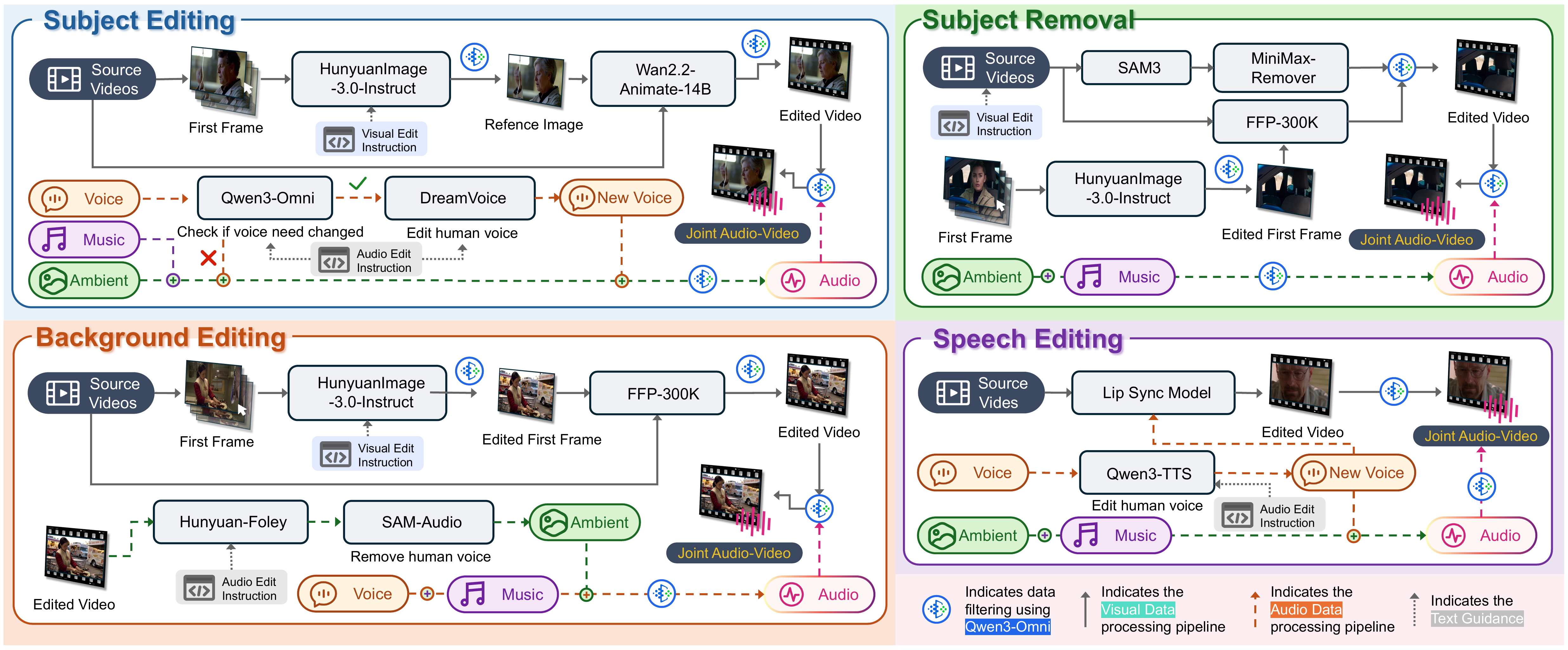}
    \vspace{-1em}
    \caption{Detailed editing pipelines of \method.
    Four dedicated pipelines, subject editing, background editing, subject removal,
    and speech editing, jointly cover the five supported editing categories,
    where subject addition shares the subject removal pipeline and is obtained by reversing
    its inputs and outputs.
    Each pipeline processes the visual and audio streams independently and
    recombines them into the final edited video. The source video frames shown in the figure are sampled from OpenHumanVid\cite{OpenHumanVid}.}
    \label{fig:pipelines}
\end{figure*}

As shown in Figure~\ref{fig:pipelines}, four dedicated pipelines cover the five editing categories:
subject editing, background editing, subject removal, and speech editing each have a dedicated pipeline,
while subject addition reuses the subject removal pipeline with reversed inputs and outputs.
Each pipeline processes visual and audio streams independently and recombines them into the final video.
Qwen3-Omni serves as an intermediate quality checker at key stages to prevent error accumulation.

\noindent\textbf{Subject Editing.}

HunyuanImage-3.0 Instruct~\cite{HunYuanImage} edits the subject appearance in a reference frame,
which drives Wan2.2-Animate~\cite{Wan22} in replace mode to generate a temporally consistent video.
On the audio side, DreamVoice~\cite{DreamVoice} converts the voice style or timbre per the instruction
while preserving spoken content, then recombines it with the original music and ambient sound.

\noindent\textbf{Background Editing.}
\label{subsubsec:bg_edit}

HunyuanImage-3.0 Instruct edits the first frame to reflect the new scene while preserving the foreground subject
without an explicit mask; FFP-300K~\cite{FFP300K} then generates a temporally consistent video from this reference.
On the audio side, HunyuanVideo-Foley~\cite{HunyuanFoley} synthesizes ambient sound conditioned on the edited video;
SAM-Audio removes any residual voice or music, and the cleaned ambient sound is recombined with the original voice and music.

\noindent\textbf{Subject Removal.}
\label{subsubsec:removal}

We employ two complementary visual routes in parallel.
MiniMax-Remover~\cite{MinimaxRemover} applies SAM3~\cite{SAM3} mask-guided inpainting, which is well suited for subjects appearing in mid-frames.
Alternatively, HunyuanImage-3.0 Instruct generates a subject-free reference frame that drives FFP-300K~\cite{FFP300K}, better handling first-frame subjects.
The higher-quality result from the two routes is selected as the final output.
On the audio side, the voice stream is discarded and the remaining music and ambient sound are recombined.
Subject Addition data is obtained by swapping the source and target of each removal triplet and rewriting the instruction accordingly.

\noindent\textbf{Speech Editing.}
\label{subsubsec:speech}

This pipeline follows an audio-first order: Qwen3-TTS~\cite{Qwen3TTS} performs zero-shot voice cloning
to synthesize new spoken content while preserving the speaker's identity and timbre;
a lip-sync model~\cite{LatentSync} then drives the source video's lip motion to match the new speech.

\subsection{Agent-in-the-loop Quality Control}
\label{subsec:agent}

\begin{figure}[tp]
    \centering
    \includegraphics[width=1.0\linewidth]{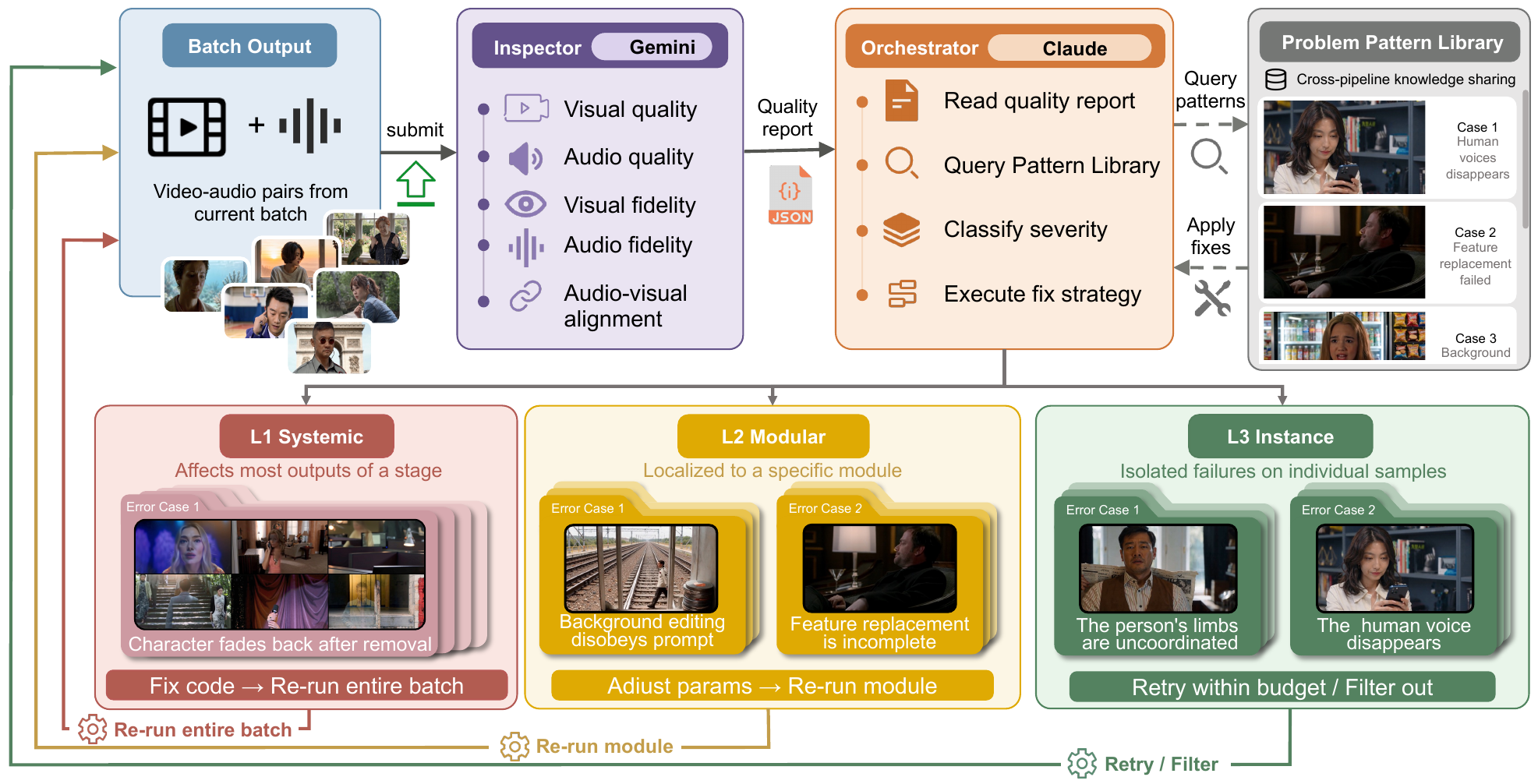}
    \vspace{-1em}
    \caption{Overview of the Agent-in-the-loop quality control framework of \textbf{\method}.
    An \textbf{Inspector} agent examines sampled outputs and produces structured quality reports,
    while an \textbf{Orchestrator} agent classifies failures into three levels and applies
    targeted fixes, with verified solutions stored in a \textbf{Problem Pattern Library} for reuse. The source video frames shown in the figure are sampled from OpenHumanVid\cite{OpenHumanVid}.}
    \label{fig:agent}
\end{figure}

Cascaded generative models inevitably produce failures, such as misaligned reference images propagating into incorrect edits or overly strict filters discarding valid data.
Manually sampling outputs, diagnosing failures, and patching code is not scalable.
We therefore propose an \emph{Agent-in-the-loop} quality control mechanism, illustrated in Figure~\ref{fig:agent}.

\noindent\textbf{Agent Architecture.}
Unlike prior approaches that use LLMs as one-shot filters to score and discard low-quality samples~\cite{InsViE-1M,Ditto}, our mechanism is a closed-loop, multi-round system that detects failures, diagnoses root causes, patches pipeline code, adjusts parameters, and stores verified fixes for cross-pipeline reuse. This shifts quality control from passive filtering to active self-repair without human intervention.
Our framework employs two specialized agents with distinct roles.
The \textbf{Orchestrator} (Claude~Opus~4.6) serves as the central controller
of the entire quality-control system: it governs the overall loop by sampling diagnostic subsets,
classifying failures, authoring code patches, coordinating retry logic,
and invoking the Inspector as needed.
The \textbf{Inspector} (Gemini~3.1~Pro~\cite{Gemini}) is called upon by the Orchestrator
to perform high-quality examination and analysis of small batches of multimodal data,
assessing the fidelity of visual edits, the quality of audio,
and the alignment between audio and video,
and returning structured quality reports for the Orchestrator to act upon.
In practice, we conduct human inspection on a 1K subset of \dataset\ and find that
applying three rounds of Agent-in-the-loop quali raises the overall qualification rate
from 36\% to 83\%.

\noindent\textbf{Hierarchical Problem Classification.}
The Orchestrator classifies detected problems into three levels.
\textbf{L1 Systemic Issues} affect the majority of outputs from a pipeline stage due to flawed
prompt templates or incorrect logic, prompting the Orchestrator to modify the pipeline code or
prompt template and re-run generation.
\textbf{L2 Modular Issues} are confined to a specific pipeline module
(e.g., a misconfigured threshold discarding excessive valid data),
and are resolved by adjusting the relevant module's parameters without touching other components.
\textbf{L3 Instance-level Issues} are isolated failures on individual samples caused by
stochastic generation artifacts, handled by retrying within a fixed budget or invoking the
Inspector to filter out defective instances.
Concrete examples of each failure level and their corresponding fixes are illustrated in Figure~\ref{fig:agent}.

\noindent\textbf{Problem Pattern Library.}
To prevent redundant re-diagnosis of recurring failure modes, we maintain a
\emph{Problem Pattern Library}, a persistent key-value store that maps problem descriptions
to verified fixes.
Before attempting to resolve a new problem, the Orchestrator first consults the library;
if a matching pattern is found, the stored fix is applied directly.
Successful fixes are added to the library, enabling knowledge sharing across pipelines
and accelerating quality control as the dataset grows.

\subsection{Dataset Statistics}
\label{subsec:dataset_stats}

The final \dataset\ dataset contains 103K high-quality joint audio-visual editing triplets
across five well-balanced categories (Figure~\ref{fig:data_overview}).
We highlight three aspects of diversity.
\emph{Linguistic diversity}: instructions span a broad vocabulary of entities and actions
in both concise and detailed forms.
\emph{Audio diversity}: each task engages distinct combinations of voice, music, and ambient sound,
and Speech Editing alone covers 32 topic domains.
\emph{Quality assurance}: SyncNet~\cite{SyncNet} scores confirm strong face-voice alignment,
and VTSS indicates reliable visual quality across all tasks.
As summarized in Table~\ref{tab:dataset_comparison}, \dataset\ is the only dataset
that jointly covers audio and visual editing with free-form natural language instructions,
providing a solid foundation for training and evaluating joint audio-visual editing models.

\section{\model\ for Joint Audio-Visual Editing}
\label{sec:model}

We adapt LTX-2.3 for the audio-visual editing task by formulating it as a
reference-conditioned denoising problem.
Given a reference video $V$ with its audio track $A$ and an editing instruction $p$,
the model generates an edited audio-visual pair under the guidance of the reference signals.

\noindent\textbf{Reference-Conditioned Input Construction.}
Let $X_v$ and $X_a$ denote the noisy latents of the target video and audio, respectively.
The video branch input is constructed by concatenating the reference latent with the noisy target
along the sequence dimension, yielding $[V;\, X_v]$, and similarly for the audio branch:
$[A;\, X_a]$.
To distinguish conditioning signals from denoising targets, we assign a timestep of $\sigma = 0$
to all reference positions ($V$ and $A$), indicating clean signals exempt from denoising,
while the target positions ($X_v$ and $X_a$) are assigned the sampled diffusion timestep $\sigma > 0$.

\noindent\textbf{Positional Encoding.}
For positional encoding, the reference and target sequences share the same RoPE coordinate space:
$\mathrm{RoPE}(V) = \mathrm{RoPE}(X_v)$ and $\mathrm{RoPE}(A) = \mathrm{RoPE}(X_a)$,
ensuring that the attention mechanism establishes precise spatial-temporal correspondences
between the reference and the generation target.

\noindent\textbf{Parameter-Efficient Fine-Tuning.}
We adopt a LoRA fine-tuning strategy, attaching LoRA adapters to the attention layers
($W_Q, W_K, W_V, W_O$) and feed-forward networks with a rank of 128.
The training objective is computed exclusively on target token positions:
\begin{equation}
    \mathcal{L} = \mathbb{E}_{\sigma,\,\epsilon}\!\left[
        \left\| f_\theta\!\left([V;\,X_v],\,[A;\,X_a],\,p,\,\sigma\right)
        - (\epsilon - X_0) \right\|^2 \cdot \mathbf{M}
    \right],
    \label{eq:loss}
\end{equation}
where $f_\theta$ denotes the model prediction, $X_0$ is the clean target,
$\epsilon$ is the sampled noise, and $\mathbf{M}$ is a binary mask that equals 1
at target token positions and 0 at reference token positions.

\section{Constructing the \benchmark\ Benchmark}
\label{sec:benchmark}

\noindent\textbf{Test Set.}
\benchmark\ consists of 150 source videos manually curated to ensure diversity
across scene type, human subject characteristics, and audio composition
spanning voice, music, and ambient sound.
Editing instructions for all five tasks are manually reviewed to guarantee quality and feasibility.
The per-task sample counts and sub-category breakdowns are detailed in Figure~\ref{fig:data_overview}.

\noindent\textbf{Evaluation Metrics.}
We construct six metrics spanning five evaluation dimensions,
combining MLLMs with traditional models.
\textbf{(1)~Visual Quality} is measured by VTSS.
\textbf{(2)~Audio Quality} is measured by UTMOSv2~\cite{UTMOSv2}.
\textbf{(3)~Audio-Visual Synchrony} is measured by SyncNet.
\textbf{(4)~Instruction Following} is assessed by Qwen3-Omni,
which scores whether the edited video faithfully executes the editing instruction.
\textbf{(5)~Video Fidelity} is also assessed by Qwen3-Omni,
which scores whether the content unrelated to the instruction is preserved.
Beyond these five dimension-specific metrics, a sixth metric employs Qwen3-Omni
to perform a holistic joint audio-visual quality assessment of the edited video.
We validate the human alignment of all six metrics
through a pairwise preference study with 5 expert annotators on 60 sampled videos,
achieving Spearman's $\rho \geq 0.80$ across all metrics
(details in Appendix~\ref{app:human_alignment}).

\section{Experiments}
\label{sec:experiments}

\subsection{Experimental Setup}
\label{subsec:exp_setup}

\noindent\textbf{Baselines.}
Because no prior work directly addresses instruction-guided joint audio-visual editing,
we compare \model\ against three representative baselines.
\textbf{(1)~AVED}~\cite{AVED} requires a source-target prompt pair as input.
\textbf{(2)~AVI-Edit}~\cite{AVIEdit} requires a segmentation mask of the editing target
together with a target prompt.
To make both methods applicable to \benchmark,
we use an LLM to automatically convert each natural-language instruction
into the corresponding input format.
\textbf{(3)~Sequential} cascades Kiwi-Edit~\cite{Kiwi-Edit},
the strongest open-source video editing model supporting 720p output,
with HunyuanVideo-Foley for video audio dubbing,
representing a strong non-joint alternative.
Inference details for all baselines are provided in the appendix.

\subsection{Main Results}
\label{subsec:main_results}

\begin{table*}[t]
\centering
\caption{Quantitative comparison on \benchmark\ across five evaluation dimensions.
Best results are \textbf{bolded}.
Sequential cascades Kiwi-Edit~\cite{Kiwi-Edit} with HunyuanVideo-Foley~\cite{HunyuanFoley}.}
\label{tab:main_results}
\resizebox{0.75\textwidth}{!}{%
\begin{tabular}{lcccccc}
\toprule
\textbf{Method} & \textbf{Visual} & \textbf{Audio} & \textbf{AV} & \textbf{Instruction} & \textbf{Video} & \textbf{AV} \\
 & \textbf{Quality} $\uparrow$ & \textbf{Quality} $\uparrow$ & \textbf{Sync} $\uparrow$ & \textbf{Compliance} $\uparrow$ & \textbf{Fidelity} $\uparrow$ & \textbf{Quality} $\uparrow$ \\
\midrule
AVED
  & 0.0590 & 1.72 & 0.1641
  & 2.95 & 3.87 & 2.93 \\
AVI-Edit
  & \textbf{0.0604} & 2.34 & 0.2721
  & 3.49 & 3.89 & 3.86 \\
Sequential
  & 0.0563 & 2.35 & 0.2925
  & 3.99 & 4.08 & 3.51 \\
\midrule
\model\ (Ours)
  & 0.0596 & \textbf{2.42} & \textbf{0.3688}
  & \textbf{4.07} & \textbf{4.22} & \textbf{3.88} \\
\bottomrule
\end{tabular}%
}
\end{table*}

\noindent\textbf{Quantitative Comparison.}

As shown in Table~\ref{tab:main_results}, \model\ ranks first on five of six metrics.
It substantially outperforms AVED and AVI-Edit on instruction compliance and audio-visual quality, further confirming the limitations of the source-target prompt paradigm discussed in Section~\ref{sec:intro}.
Compared with Sequential, joint modeling yields a 26\% relative gain on audio-visual synchrony, as cascading independent editors inevitably introduces cross-modal misalignment.
AVI-Edit holds a marginal lead on Visual Quality owing to its explicit mask that constrains edits to a localized region.
Per-task breakdowns are provided in Appendix~\ref{app:per_task}.

\begin{figure*}[t]
    \centering
    \includegraphics[width=1.0\linewidth]{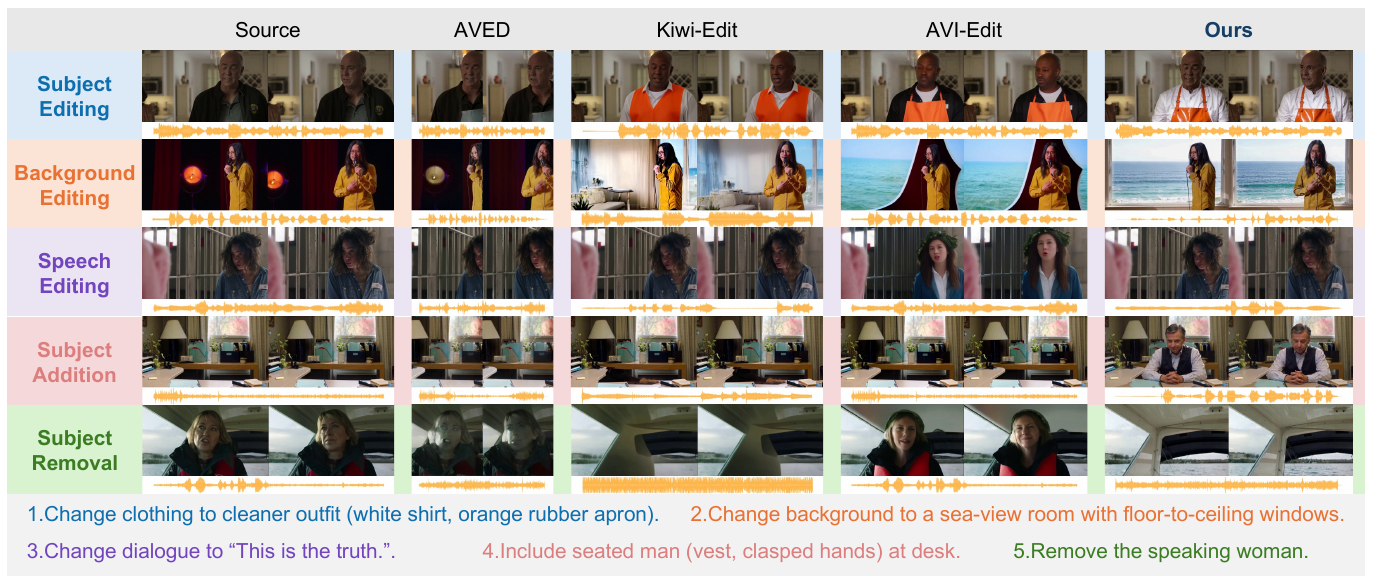}
    \vspace{-1em}
    \caption{\textbf{Qualitative comparison on \benchmark.}
    Rows show outputs of the source video and each method; columns correspond to the five editing task categories. The source video frames shown in the figure are sampled from OpenHumanVid\cite{OpenHumanVid}.}
    \label{fig:qualitative}
\end{figure*}

\noindent\textbf{Qualitative Comparison.}
Figure~\ref{fig:qualitative} presents visual comparisons across all five editing tasks.
AVED and AVI-Edit tend to produce over-smoothed or semantically inconsistent results due to their reliance on automatically converted prompts.
Sequential maintains reasonable visual quality but suffers from audio-visual misalignment since the audio dubbing module operates without awareness of the visual edits.
\model\ consistently generates edits that are visually coherent, semantically faithful to the instruction, and temporally synchronized across both modalities.
Additional per-task comparisons are provided in Appendix~\ref{app:qualitative}.

\subsection{Ablation Study, Analysis, and Observation}
\label{subsec:ablation}

\noindent\textbf{Effect of Agent-in-the-loop Quality Control.}

To isolate the contribution of the Agent-in-the-loop quality control,
we construct a control dataset of the same scale as \dataset\
but produced \emph{without} any agent intervention,
and fine-tune LTX-2.3 on it under identical configurations to obtain \model\ w/o Agent.
We also vary the training set size across three scales (5K, 15K, 100K) to study data scaling.
Both ablations are reported together in Table~\ref{tab:ablation}.

\noindent\textbf{Effect of Data Scale.}
Performance improves consistently as training data grows,
with the largest gains observed between 5K and 15K,
and continued but diminishing improvements from 15K to 100K.

\begin{table}[h]
\centering
\caption{Ablation study on \benchmark.
\model-tiny and \model-small are trained on 5K and 15K samples, respectively;
\model\ (Ours) uses the full 100K.
Row 4 removes Agent-in-the-loop QC at full scale.
All models are fine-tuned from LTX-2.3 under identical configurations.
Best results are \textbf{bolded}.}
\label{tab:ablation}
\resizebox{0.9\columnwidth}{!}{%
\begin{tabular}{llccccccc}
\toprule
\textbf{Model} & \textbf{Scale} & \textbf{Agent} & \textbf{Visual} & \textbf{Audio} & \textbf{AV} & \textbf{Instruction} & \textbf{Video} & \textbf{AV} \\
 & & \textbf{QC} & \textbf{Quality}$\uparrow$ & \textbf{Quality}$\uparrow$ & \textbf{Sync}$\uparrow$ & \textbf{Compliance}$\uparrow$ & \textbf{Fidelity}$\uparrow$ & \textbf{Quality}$\uparrow$ \\
\midrule
\model-tiny & 5K   & \checkmark & 0.0574 & 2.38 & 0.2453 & 3.21 & 3.95 & 3.52 \\
\model-small & 15K  & \checkmark & 0.0579 & \textbf{2.44} & 0.2871 & 3.49 & \underline{4.18} & \underline{3.84} \\
\model\ w/o Agent & 100K & $\times$ & \underline{0.0581} & 2.31 & \underline{0.3012} & \underline{3.61} & 4.05 & 3.63 \\
\model\ (Ours) & 100K & \checkmark & \textbf{0.0596} & \underline{2.42} & \textbf{0.3688} & \textbf{4.07} & \textbf{4.22} & \textbf{3.88} \\
\bottomrule
\end{tabular}}
\end{table}

\noindent\textbf{Analysis.}
Together, the two ablations reveal complementary scaling axes.
Agent-in-the-loop quality control acts as a \emph{quality} filter:
even at fixed dataset size, removing it causes a consistent drop across all \benchmark\ dimensions,
indicating that noisy training pairs hurt audio-visual alignment more than they help through sheer volume.
Data scale, on the other hand, acts as a \emph{quantity} driver:
performance improves consistently as training data grows from 5K to 100K,
confirming that a larger pool of quality-controlled data translates directly into stronger editing capability.
These findings jointly justify both the scale of \dataset\ and the design of our agent-driven curation pipeline.

\section{Conclusion}
\label{sec:conclusion}

In this work, we present \dataset, a dataset for instruction-guided joint audio-visual editing. To address the scalability and alignment challenges in cross-modal data synthesis, we introduce an automated Agent-in-the-loop pipeline. This mechanism replaces manual curation to maintain spatiotemporal and semantic coherence at scale. Furthermore, we establish \benchmark\ to evaluate structural and speech-related edits. \model\ achieves state-of-the-art performance on \benchmark.

\noindent\textbf{Limitations and Future Work.} While the Agent-in-the-loop pipeline significantly mitigates cross-stage error accumulation, the success rate on highly complex editing tasks involving multiple simultaneous changes remains limited when the capability of the underlying foundation models is fixed. Additionally, our current dataset primarily focuses on human-centric scenarios. Future work will extend this paradigm to open-domain audio-visual environments and explore more capable base models to further improve editing quality on challenging cases.

\bibliography{references}

\medskip

\appendix

\newpage
\renewcommand{\thesection}{\Alph{section}}
\setcounter{section}{0}

\begin{center}
    \textbf{\Large \method: Joint Audio-Visual Instruction-Guided}\\[0.5em]
    \textbf{\Large Video Editing with Agentic Data Curation}\\[1.8em]
    \large Supplementary Material
\end{center}

\vspace{1.6em}
{\large \textbf{Contents}}

\startcontents[appendices]
\printcontents[appendices]{l}{1}{\setcounter{tocdepth}{3}}

\section{Related Work}
\label{sec:related_work}

\subsection{Instruction-Guided Audio-Visual Editing Datasets}
\label{subsec:rw_datasets}

Existing large-scale instruction-guided video editing datasets~\cite{InsViE-1M,Ditto,OpenVE-3M} focus exclusively on visual transformations, providing no paired audio-visual editing examples in which both modalities are jointly modified under a single natural-language instruction. While recent model-in-the-loop pipelines have improved data scale and diversity, they still rely on human inspection to diagnose failures and patch pipeline code, which fundamentally limits scalability. LLM-based agents have been applied to automate machine learning research~\cite{MetaClaw,AutoResearchRL}, but not to data pipeline quality control. \dataset\ addresses both gaps as the first large-scale dataset for instruction-guided joint audio-visual editing, constructed via an \emph{Agent-in-the-loop} pipeline that autonomously performs hierarchical quality diagnosis and code-level self-repair.

\subsection{Instruction-Guided Audio-Visual Editing Methods}
\label{subsec:rw_methods}

Driven by the rapid advancement of generative models\cite{ho2020denoising,peebles2023scalable,chen2025dip,ho2022classifier,chen2025ragd,song2020score,chen2026l2p}, instruction-guided video editing models~\cite{InsViE-1M,Ditto,OpenVE-3M} have gained significant momentum, yet they consistently overlook audio as an essential modality in video. The few existing joint audio-visual editing works, AVED~\cite{AVED}, AV-Edit~\cite{AVEdit}, and AVIEdit~\cite{AVIEdit}, follow a \emph{source-target prompt-based} paradigm that requires users to supply a pair of full captions rather than a natural-language instruction, and are largely confined to attribute-level modifications, failing to support structural edits such as subject removal or speech editing. \model\ is trained end-to-end on \dataset\ and natively accepts free-form instructions across five editing categories, to our knowledge the first instruction-guided joint audio-visual editing model.

\subsection{Instruction-Guided Audio-Visual Editing Benchmarks}
\label{subsec:rw_benchmarks}

Existing video editing benchmarks~\cite{InsViE-1M,OpenVE-3M} evaluate the visual stream only, relying on metrics such as CLIP score, PSNR, or VLM-based frame assessment, with no mechanism for measuring whether the audio track has been appropriately modified. \benchmark\ fills this gap by jointly evaluating visual-audio quality, instruction compliance, and video fidelity, providing the first comprehensive evaluation protocol for instruction-guided joint audio-visual editing.

\section{Experimental Details}
\label{app:exp_details}

\noindent\textbf{Baseline Inference Details.}

For \textbf{AVED}~\cite{AVED}, we use the detailed caption of the source video as the
source prompt, and employ Qwen3-Omni to generate the target prompt
by combining the video caption with the editing instruction.
The output resolution follows the official default of $512{\times}512$.
For \textbf{AVI-Edit}~\cite{AVIEdit}, we use Qwen3-Omni to extract a textual description
of the editing target from the editing instruction, and then follow the official default
configuration to obtain the mask video using SAM2~\cite{SAM2} conditioned on this description.
The output resolution follows the official default of $1280{\times}736$.
For the \textbf{Sequential} baseline, Kiwi-Edit~\cite{Kiwi-Edit} produces the edited video
at $1280{\times}720$, after which HunyuanVideo-Foley takes the
edited video as the conditioning input for audio dubbing.
All model parameters follow their respective official default configurations.

\noindent\textbf{Training Details.}

We adopt a LoRA fine-tuning strategy on LTX-2.3, attaching LoRA adapters to the attention layers ($W_Q, W_K, W_V, W_O$) and feed-forward networks with a rank of 128.

\section{Human Alignment of Evaluation Metrics}
\label{app:human_alignment}

To validate that the automatic metrics employed in \benchmark\ align with human judgment,
we conduct a human evaluation study following established protocols~\cite{IVEBench}.

\noindent\textbf{Study Design.}
We randomly sample 60 source videos from the \benchmark\ test set (12 per editing task)
and collect the edited outputs from all four methods (AVED, AVI-Edit, Sequential, and \model),
yielding $60 \times 4 = 240$ edited videos in total.
For each source video, $\binom{4}{2} = 6$ pairwise comparisons are constructed,
resulting in $60 \times 6 = 360$ pairwise evaluation instances.

\noindent\textbf{Annotators and Protocol.}
We recruit 5 annotators with professional backgrounds in video production or computer vision research.
All annotators undergo a calibration session with 10 practice examples and detailed scoring guidelines
before the formal evaluation.
For each pairwise comparison, annotators are presented with the source video, the editing instruction,
and two edited videos (order randomized), and are asked to judge which video performs better
along three dimensions: (1) Instruction Compliance, (2) Video Fidelity, and (3) Overall AV Quality.
Annotators may select ``hard to distinguish'' if the two videos are of comparable quality.
Scores are assigned as 1.0 for the preferred video, 0.0 for the other, and 0.5 for ties.
The final human preference score for each method--dimension pair is obtained by averaging
across all annotators and all relevant pairwise comparisons.

\noindent\textbf{Inter-Annotator Agreement.}
We compute Fleiss' $\kappa$ to measure inter-annotator agreement.
The obtained values are $\kappa = 0.72$ for Instruction Compliance,
$\kappa = 0.68$ for Video Fidelity, and $\kappa = 0.74$ for Overall AV Quality,
all indicating substantial agreement ($\kappa > 0.6$).

\noindent\textbf{Correlation with Automatic Metrics.}
Table~\ref{tab:human_alignment} reports Spearman's rank correlation coefficient ($\rho$)
between the human preference rankings and the automatic metric rankings across the four methods.
All three MLLM-based metrics (Instruction Compliance, Video Fidelity, AV Quality scored by Qwen3-Omni)
achieve high correlation with human judgment ($\rho \geq 0.90$),
validating their reliability as evaluation proxies.
The traditional metrics (VTSS, UTMOSv2, SyncNet) also demonstrate moderate-to-strong correlations,
confirming that the full metric suite of \benchmark\ is well-aligned with human perception.

\begin{table}[H]
\centering
\caption{Spearman's rank correlation ($\rho$) between automatic metrics and human preferences.
Human evaluation is conducted on 60 sampled videos across all five tasks with 5 annotators.}
\label{tab:human_alignment}
\resizebox{0.7\columnwidth}{!}{%
\begin{tabular}{lcc}
\toprule
\textbf{Automatic Metric} & \textbf{Evaluation Dimension} & \textbf{Spearman's $\rho$} \\
\midrule
Qwen3-Omni: Instruction Compliance & Instruction Following & 0.94 \\
Qwen3-Omni: Video Fidelity & Content Preservation & 0.90 \\
Qwen3-Omni: AV Quality & Overall Quality & 0.92 \\
\midrule
VTSS~\cite{Koala36M} & Visual Quality & 0.80 \\
UTMOSv2~\cite{UTMOSv2} & Audio Quality & 0.85 \\
SyncNet~\cite{LatentSync} & Audio-Visual Synchrony & 0.88 \\
\bottomrule
\end{tabular}%
}
\end{table}

\noindent\textbf{Discussion.}
The high correlations validate our choice of Qwen3-Omni as the primary MLLM judge:
its open-source nature ensures full reproducibility, and its multimodal capabilities
(joint video and audio understanding) make it uniquely suited for evaluating
audio-visual editing quality.
We note that the traditional metrics exhibit slightly lower but still strong correlations,
suggesting they capture complementary signal-level information (e.g., perceptual audio quality, pixel-level synchrony)
that MLLM-based holistic scoring may occasionally overlook.
Together, the six metrics provide a comprehensive and human-aligned evaluation framework for \benchmark.

\section{Per-Task Breakdown on \benchmark}
\label{app:per_task}

Tables~\ref{tab:app_utmos}--\ref{tab:app_avquality} report per-task scores for all six evaluation metrics on \benchmark.
Task abbreviations: \textbf{Sub.Edit} = subject editing, \textbf{BG.Edit} = background editing,
\textbf{Speech} = speech content modification, \textbf{Sub.Add} = subject addition, \textbf{Sub.Rm} = subject removal.

\begin{table}[H]
\centering
\caption{Audio Quality per task (higher is better).}
\label{tab:app_utmos}
\resizebox{0.75\columnwidth}{!}{%
\begin{tabular}{lccccccc}
\toprule
\textbf{Method} & \textbf{Overall} & \textbf{Sub.Edit} & \textbf{BG.Edit} & \textbf{Speech} & \textbf{Sub.Add} & \textbf{Sub.Rm} \\
\midrule
AVED                        & 1.72 & 1.71 & 1.49 & 1.96 & 1.62 & 1.75 \\
AVI-Edit                    & 2.34 & 2.43 & 2.28 & 2.33 & 2.17 & 2.44 \\
Sequential                  & 2.35 & 2.23 & \textbf{2.31} & 2.35 & 2.36 & \textbf{2.56} \\
\model\ (Ours)              & \textbf{2.42} & \textbf{2.48} & 2.23 & \textbf{2.62} & \textbf{2.48} & 2.24 \\
\bottomrule
\end{tabular}%
}
\end{table}

\begin{table}[H]
\centering
\caption{Visual Quality per task (higher is better).}
\label{tab:app_vtss}
\resizebox{0.75\columnwidth}{!}{%
\begin{tabular}{lccccccc}
\toprule
\textbf{Method} & \textbf{Overall} & \textbf{Sub.Edit} & \textbf{BG.Edit} & \textbf{Speech} & \textbf{Sub.Add} & \textbf{Sub.Rm} \\
\midrule
AVED                        & 0.0590 & 0.0603 & 0.0612 & 0.0603 & 0.0551 & 0.0562 \\
AVI-Edit                    & \textbf{0.0604} & \textbf{0.0625} & \textbf{0.0650} & 0.0599 & 0.0492 & \textbf{0.0638} \\
Sequential                  & 0.0563 & 0.0591 & 0.0601 & 0.0616 & 0.0564 & 0.0386 \\
\model\ (Ours)              & 0.0596 & 0.0609 & 0.0636 & \textbf{0.0625} & \textbf{0.0633} & 0.0447 \\
\bottomrule
\end{tabular}%
}
\end{table}

\begin{table}[H]
\centering
\caption{AV Sync per task (higher is better).}
\label{tab:app_sync}
\resizebox{0.75\columnwidth}{!}{%
\begin{tabular}{lccccccc}
\toprule
\textbf{Method} & \textbf{Overall} & \textbf{Sub.Edit} & \textbf{BG.Edit} & \textbf{Speech} & \textbf{Sub.Add} & \textbf{Sub.Rm} \\
\midrule
AVED                        & 0.1641 & 0.1333 & 0.0613 & 0.1050 & \textbf{0.5459} & 0.1367 \\
AVI-Edit                    & 0.2721 & 0.3236 & \textbf{0.3584} & 0.2453 & 0.0365 & \textbf{0.2658} \\
Sequential                  & 0.2925 & 0.3463 & 0.2923 & 0.2769 & 0.2497 & 0.0182 \\
\model\ (Ours)              & \textbf{0.3688} & \textbf{0.3779} & 0.3097 & \textbf{0.4569} & 0.3264 & 0.0146 \\
\bottomrule
\end{tabular}%
}
\end{table}

\begin{table}[H]
\centering
\caption{Instruction Compliance per task, scored 1 to 5, higher is better.}
\label{tab:app_compliance}
\resizebox{0.75\columnwidth}{!}{%
\begin{tabular}{lccccccc}
\toprule
\textbf{Method} & \textbf{Overall} & \textbf{Sub.Edit} & \textbf{BG.Edit} & \textbf{Speech} & \textbf{Sub.Add} & \textbf{Sub.Rm} \\
\midrule
AVED                        & 2.95 & 2.60 & 2.87 & 3.67 & 2.20 & 3.50 \\
AVI-Edit                    & 3.49 & 3.64 & 3.63 & 3.30 & 3.20 & 3.58 \\
Sequential                  & 3.99 & 4.17 & \textbf{3.97} & 3.67 & 3.84 & \textbf{4.25} \\
\model\ (Ours)              & \textbf{4.07} & \textbf{4.22} & 3.83 & \textbf{4.33} & \textbf{3.88} & 3.96 \\
\bottomrule
\end{tabular}%
}
\end{table}

\begin{table}[H]
\centering
\caption{Video Fidelity per task, scored 1 to 5, higher is better.}
\label{tab:app_fidelity}
\resizebox{0.75\columnwidth}{!}{%
\begin{tabular}{lccccccc}
\toprule
\textbf{Method} & \textbf{Overall} & \textbf{Sub.Edit} & \textbf{BG.Edit} & \textbf{Speech} & \textbf{Sub.Add} & \textbf{Sub.Rm} \\
\midrule
AVED                        & 3.87 & 3.92 & \textbf{4.07} & 4.00 & 3.48 & 3.79 \\
AVI-Edit                    & 3.89 & 4.10 & 3.93 & 3.70 & 3.72 & 3.88 \\
Sequential                  & 4.08 & 4.12 & \textbf{4.07} & 3.90 & 4.00 & \textbf{4.33} \\
\model\ (Ours)              & \textbf{4.22} & \textbf{4.30} & \textbf{4.07} & \textbf{4.33} & \textbf{4.04} & \textbf{4.30} \\
\bottomrule
\end{tabular}%
}
\end{table}

\begin{table}[H]
\centering
\caption{AV Quality per task, scored 1 to 5, higher is better.}
\label{tab:app_avquality}
\resizebox{0.75\columnwidth}{!}{%
\begin{tabular}{lccccccc}
\toprule
\textbf{Method} & \textbf{Overall} & \textbf{Sub.Edit} & \textbf{BG.Edit} & \textbf{Speech} & \textbf{Sub.Add} & \textbf{Sub.Rm} \\
\midrule
AVED                        & 2.93 & 3.10 & 3.13 & 3.73 & 1.72 & 2.68 \\
AVI-Edit                    & 3.86 & 4.03 & 3.93 & 3.87 & 3.40 & \textbf{3.96} \\
Sequential                  & 3.51 & 3.65 & 3.43 & \textbf{3.93} & 3.84 & 2.56 \\
\model\ (Ours)              & \textbf{3.88} & \textbf{4.08} & \textbf{4.07} & 3.63 & 3.72 & 3.82 \\
\bottomrule
\end{tabular}%
}
\end{table}

\section{Additional Qualitative Results}
\label{app:qualitative}

\begin{figure*}[h]
    \centering
    \includegraphics[width=1.0\linewidth]{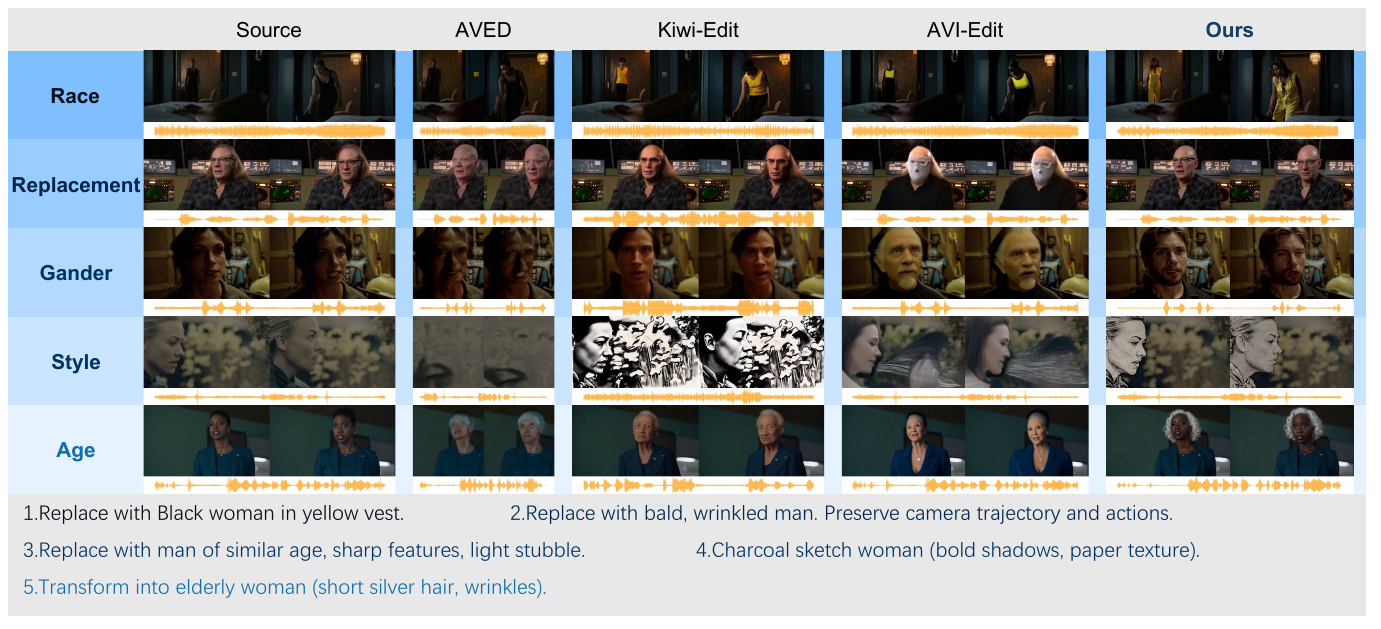}
    \caption{\textbf{Per-task qualitative comparison on subject editing.}
    Each column corresponds to a sub-task of subject editing; rows show the source video and outputs of each method. The source video frames shown in the figure are sampled from OpenHumanVid\cite{OpenHumanVid}.}
    \label{fig:app_subject}
\end{figure*}

\begin{figure*}[h]
    \centering
    \includegraphics[width=1.0\linewidth]{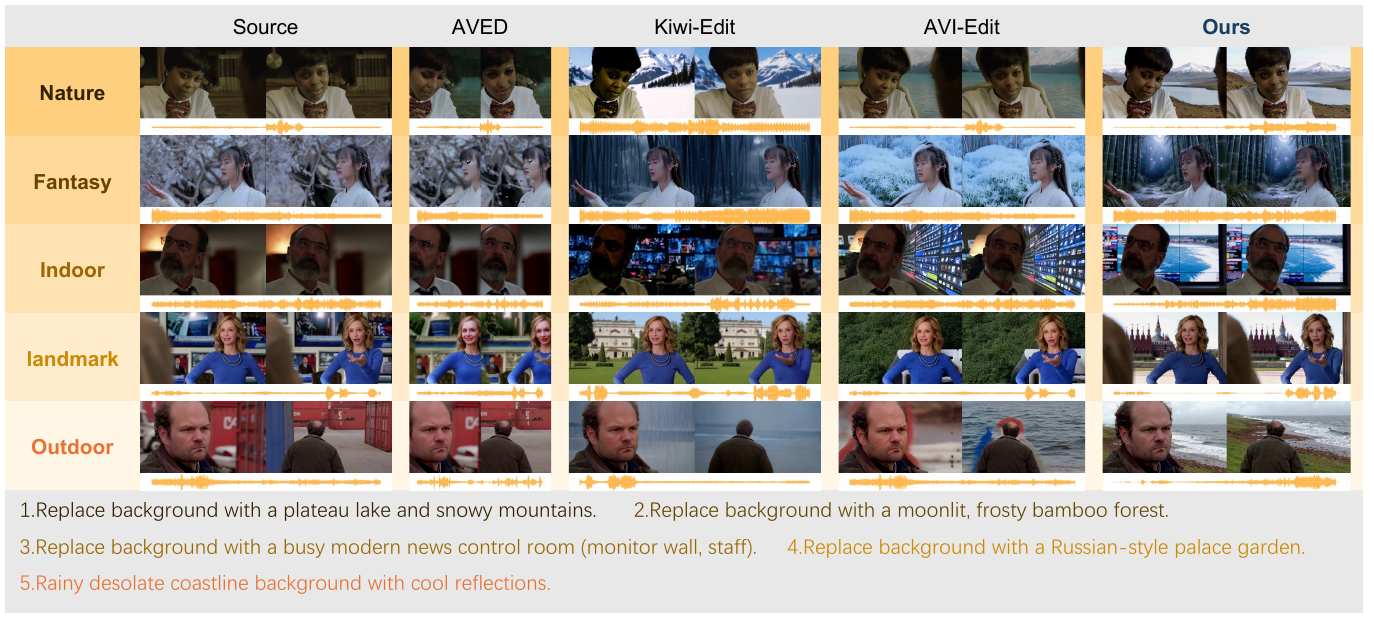}
    \caption{\textbf{Per-task qualitative comparison on background editing.}
    Each column corresponds to a sub-task of background editing; rows show the source video and outputs of each method. The source video frames shown in the figure are sampled from OpenHumanVid\cite{OpenHumanVid}.}
    \label{fig:app_background}
\end{figure*}

Figures~\ref{fig:app_subject} and~\ref{fig:app_background} provide per-task qualitative comparisons for subject editing and background editing, respectively.
Each row presents representative frames from the source video alongside the outputs of AVED, AVI-Edit, Sequential, and \model.

\clearpage

\end{document}